\documentclass[letterpaper, 10 pt, conference]{ieeeconf}  

\usepackage{graphicx}
\graphicspath{{figs/}}
\usepackage{tabularx}
\usepackage{makecell}
\usepackage{array}
\usepackage{booktabs}   
\usepackage{multirow}   
\usepackage{xcolor}
\usepackage{amsmath}
\usepackage{pifont}
\usepackage{float}
\usepackage{wrapfig}
\usepackage{pifont}
\usepackage{amssymb}
\usepackage{mathtools}
\usepackage{url}
\usepackage{threeparttable}
\usepackage{adjustbox}  
\usepackage{graphicx}   

\definecolor{navy}{RGB}{70,130,180}

\IEEEoverridecommandlockouts                              

\title{\LARGE \bf
\texttt{Sym2Real}: Symbolic Dynamics with Residual Learning \\
for Data-Efficient Adaptive Control
}
\author{Easop Lee$^{1}$, Samuel A. Moore$^{1}$ and Boyuan Chen$^{1}$
\thanks{$^{1}$ All authors are with
Duke University, Durham, NC 27708 USA. This work was supported by ARO under award W911NF2410405 and DARPA TIAMAT program under award HR00112490419.}
}

\begin{document}

\maketitle
\thispagestyle{empty}
\pagestyle{empty}


\begin{abstract}

We present \texttt{Sym2Real}, a fully data-driven framework for highly data-efficient adaptation of low-level controllers. 
Although symbolic regression is data-efficient, its role in real-world control has been limited due to its sensitivity to measurement noise, which corrupts the equations and leads to model degradation when fitted directly on real-world data.
\texttt{Sym2Real} addresses this limitation by 1) learning first from low-fidelity simulation, where noise-free trajectories allow symbolic regression to identify the underlying dynamics, and 2) using a small amount of real-world data for targeted residual adaptation to bridge the sim-to-real gap.
Using only $\sim$10 trajectories, we achieve robust control of both a quadrotor and a racecar in the real world, without expert knowledge or simulation tuning.
Through experimental validation on both platforms, we demonstrate consistent data-efficient adaptation across 6 out-of-distribution sim2sim scenarios and successful sim2real transfer across 5 real-world conditions.
More information can be found at 
\textcolor{navy}
{\url{http://generalroboticslab.com/Sym2Real}}.
\end{abstract}

\section{Introduction}

Once assembled, a robot must rapidly learn the low-level skills needed to move and act.
Beyond 
initial control, it must also continuously adapt to out-of-distribution (OOD) conditions by efficiently refining its controller rather than undergoing complete retraining, avoiding lengthy update periods that take it out of service. 
This efficiency across the robot's entire 
life cycle, from rapid early development to flexible adaptation, remains unsolved, limiting robots from achieving physical intelligence seen in biological systems.

Data-driven approaches offer promise for such adaptive control and have enabled impressive real-world performance across many robotic platforms, from quadrotors \cite{kaufmann2023champion} to autonomous vehicles \cite{8957584} and legged robots \cite{lee2020learning}.
However, at their core, these successes often rely on large, high-quality datasets that can accurately reflect target real-world dynamics. 
Collecting this data continues to be a bottleneck, no matter whether it is done in simulation or in the real world. 
Real-world data collection requires custom teleoperation interfaces, safety management, and long hours of supervision.
Simulators, while seemingly more scalable, are no less demanding.
A robot model and physics engine merely serve as a starting point; achieving high-fidelity behavior typically requires identifying and injecting task-specific and hardware-specific priors such as actuator delay, sensor noise, and external disturbances. 
To further improve transferability, domain randomization \cite{tobin2017domain} is commonly added, but selecting effective parameters and their distributions remains a trial-and-error process that presupposes dynamics knowledge in the target domain, limiting reusability across platforms or tasks \cite{ 11017653, gronauer2022using}.
Together, these challenges collectively impede the development of robust and adaptive low-level data-driven controllers, leading us to a key question: \textit{Can we develop a method that adapts across domain shifts, including and beyond sim2real gap, without large datasets, heavy simulation engineering, or prior system knowledge?}

\begin{figure}[t]
\includegraphics[width=\linewidth]
{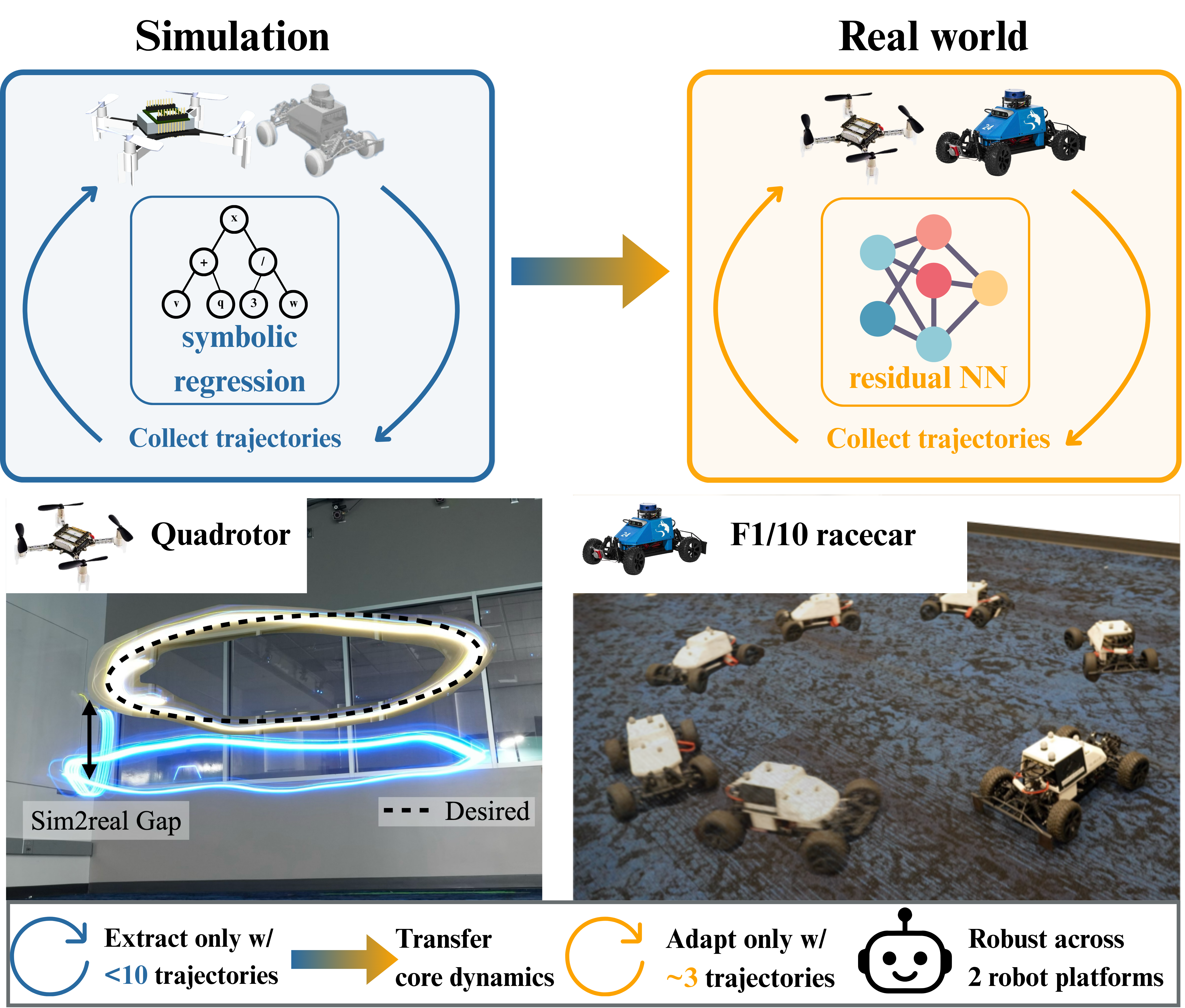}
    \caption{\textbf{\texttt{Sym2Real}} enables robust and adaptive real-world control with a small amount of simulation and real-world data.}
    \label{fig:teaser}
    \vspace{-10pt}
\end{figure}

To tackle the data-efficiency challenge, we adopt a model-based approach. However, model learning can still be data-intensive. 
A single model tasked to learn everything, from core physics to subtle differences for adaptation, overwhelms conventional learning approaches. Instead, we decompose the modeling problem with the intuition that each robot type obeys shared underlying physics, independent of environment or embodiment-specific variations. 
Rather than relying on an all-encompassing model, we propose a structured hybrid strategy: capture the shared physics with a compact, data-efficient model from low-fidelity simulation, then adapt domain-specific effects with a lightweight residual trained on minimal target data. 

Low-fidelity simulation offers safe and inexpensive data collection. Fortunately, the underlying physics principles we hope to extract are largely contained even when there is a significant domain gap with the physical world. 
Our approach does not rely on the simulation being accurate, but on its ability to provide clean, noise-free trajectories for structure identification.
We propose symbolic regression (SR) to extract the core structure, as it yields compact equations, data-efficiently. Noiseless, clean data enhances SR’s ability, turning the low-fidelity setting into an advantage. Previous work has demonstrated success with SR in simple domains like CartPole and 2D vehicles in simulation \cite{kamienny2022symbolicmodelbased, 10801405}, but its use in \emph{real-world robotics} under \emph{domain shifts} remains limited. 
We experimentally find that directly fitting SR models to real-world data often fails in practice due to measurement noise, 
resulting in structurally incorrect expressions that are physically inconsistent and unsafe for deployment.
This highlights the need for a structured, simulation-informed approach when applying SR in real-world settings.

In this work, we introduce \texttt{Sym2Real} (Fig.~\ref{fig:teaser}), a fully data-driven pipeline for learning adaptive controllers using minimal data and no simulation tuning. Our method efficiently transfers robot description files to physical control in $\sim$10 trajectories.
Across various OOD scenarios on quadrotor and racecar platforms, our method achieves consistent sim2sim and sim2real transfer. 
It extracts core physics with $<20$ simulation trajectories and enables few-shot adaptation with $<10$ trajectories across 6 simulation scenarios. In real-world tests, we adapt in $<5$ trajectories across 5 scenarios.
Our experiments also reveal a principled functional decomposition: symbolic regression is most effective for identifying shared structural dynamics from clean simulation data, while lightweight neural networks (NNs) efficiently capture domain-specific residual effects during adaptation.
\section{Related Work}

\subsection{Data Efficient Dynamics Modeling for Control} 
In model-based control, model choices often involve a trade-off between sample efficiency and generalization versus expressiveness and accuracy. 
While NN models offer sufficient expressiveness to approximate complex dynamics involving chaotic systems \cite{Moore2025Automated}, contact forces \cite{chua2018deep}
and ground effect \cite{lambert2019low},
they require large datasets to cover 
enough of the input space and generalize poorly beyond the seen samples. 

To address these limitations, a range of alternative data-driven models has been explored for model-based robot control, 
including Gaussian processes \cite{10.5555/3104482.3104541, nagabandi2018neural}, sparse dictionary learning \cite{Zolman2025Sindy}, 
physics-informed NNs \cite{gu2024physics, liu2021physics}, and symbolic regression \cite{kamienny2022symbolicmodelbased, 10801405}. 
While these methods consistently demonstrate improved generalization and sample efficiency in simulation, few have been extended to real-world settings, highlighting an opportunity to leverage such models for transfer and adaptation in robotics.

In particular, SR shows superior sample efficiency and generalization by discovering interpretable mathematical expressions that best fit data. Seminal work used evolutionary algorithms \cite{889734, schmidt2009distilling}, while recent work incorporates deep learning to improve expressiveness and search efficiency \cite{
petersen2021deep}. Open-source libraries \cite{cranmer2023interpretable, 
scherk2025symmatikastructureawaresymbolicdiscovery} have lowered the barrier to applying SR in practical settings.
In robotics, SR has been applied to dynamics modeling \cite{kamienny2022symbolicmodelbased, 10801405} and policy learning \cite{acero2024distilling}  
though only in simulation. 
Our work extends SR to real-world control by leveraging a low-fidelity simulation and sampling-based trajectory optimization, addressing the key challenges that limit direct application to physical robots.

\subsection{Real-World Control Adaptation}
Residual approaches efficiently refine existing analytical or data-driven models with only small amounts of data from the target domain.
Various types of models have been employed for residual learning in the real world, including Gaussian processes for aerial robots \cite{4209179, torrente2021data} 
and mobile vehicles \cite{
ning2023scalable}, neural ordinary differential equations for quadrotors \cite{chee2022knode} and hexacopters \cite{djeumou2023how}, and black-box NNs for soft robots \cite{gao2024sim} and quadrupeds \cite{levy2024learning}. Some have further integrated residual models with reinforcement learning policies \cite{huang2023datt}. The residuals can be learned at a much lower expense of data, naturally enabling few-shot learning. Prior work demonstrates achieving robust control using only minutes of real-world data for quadruped locomotion \cite{levy2024learning}, less than a few seconds of data for quadrotor control under aerodynamic effects \cite{torrente2021data, pan2025learning}, and minutes of data for operation under extreme wind \cite{o2022neural}. 

Many residual approaches rely on predefined nominal equations derived from first principles.
Our approach relaxes this requirement by learning symbolic models directly from simulation. While simulators are constructed from underlying physics-based equations, our method does not assume access to those equations themselves. 
Instead, we only require data generated through simulation runs, treating the simulator as a black-box oracle. 
This shift simplifies the pipeline and turns the whole framework fully data-driven.
\begin{figure*}[t]
    \vspace{5pt}
    \centering
    \includegraphics[width=\textwidth]{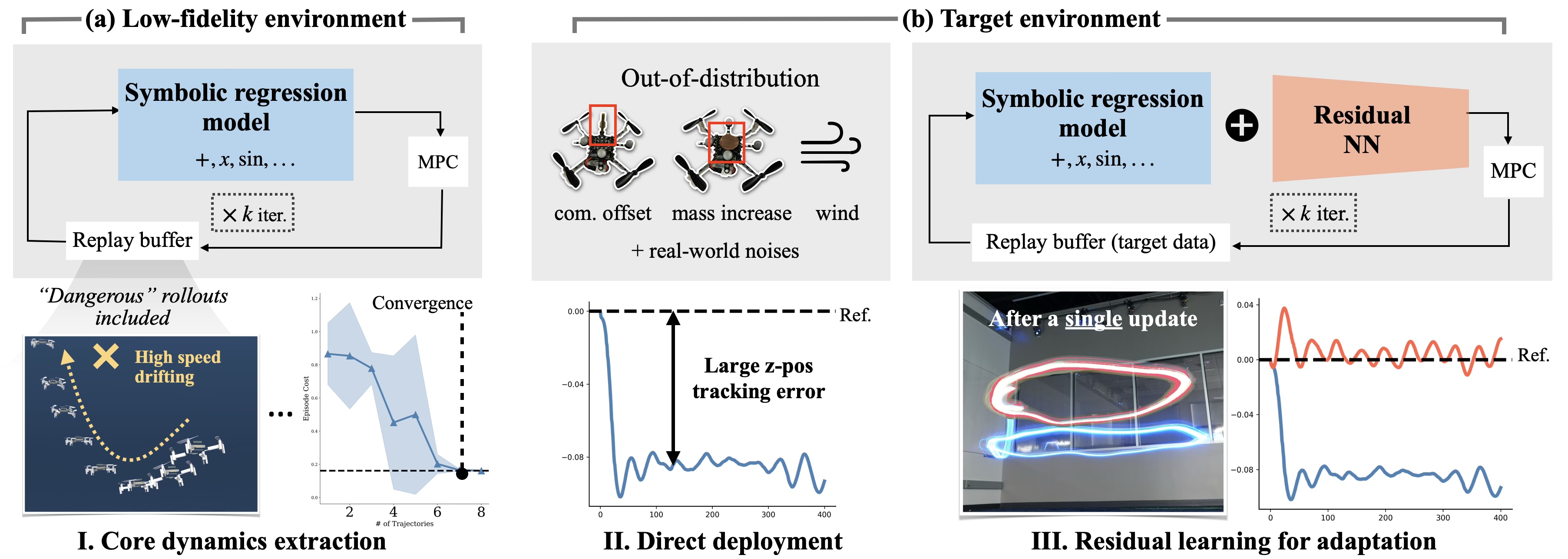}
    \caption{\textbf{\texttt{Sym2Real}} achieves data-efficient adaptive control through a two-stage process.
    (a) Simplified Environment: In Stage I, the SR dynamics model extracts core equations from a simplified low-fidelity environment, where trajectories, including high-speed drifts or flips that would trigger safety shutdowns in reality, can be safely simulated to provide rich state–action coverage from only a few runs.
    (b) Target Environment: In Stage II, directly deploying the SR model to an unseen environment leads to degraded tracking under disturbances such as noise, mass shifts, or wind. In Stage III, a small amount of target data trains a residual neural network on top of the SR equation that adapts and specializes the model to the new setting.
    }
    \label{fig:method}
    \vspace{-10pt}
\end{figure*}

\section{Preliminary}\label{preliminary}
To ground our approach, we briefly outline the iterative learning‐based model predictive control framework. 
Intuitively, it alternates between two stages: 1) model fitting and 2) data collection via model predictive control (MPC). This iteration is repeated until the task performance is satisfactory.

In this work, we define the \textit{model} as a discrete single-step dynamics function, $f: \mathcal{S} \times \mathcal{A} \rightarrow \mathcal{S}$, that maps the current state, $\mathbf{s}_t \in \mathcal{S}$, and action, $\mathbf{a}_t \in \mathcal{A} $, to the next state $\mathbf{s}_{t+1}$:
\begin{equation}
\mathbf{s}_{t+1} = f(\mathbf{s}_t, \mathbf{a}_t)
\end{equation}
The state $\mathbf{s}_t$ includes robot observations such as pose and velocity, while $\mathbf{a}_t$ represents the robot’s control input. The learned dynamics model can be queried in an auto-regressive manner to simulate future trajectories.

Given \( f \), we deploy it with the Model Predictive Path Integral (MPPI) controller \cite{7487277}. 
At each timestep, MPPI samples \( N \) action sequences over a horizon \( H \) and simulates resulting trajectories with \( f \). 
Each trajectory is evaluated using a known cost function \( c: \mathcal{S} \times \mathcal{A} \rightarrow \mathbb{R} \), which encodes task-specific objectives:
\begin{equation}
    J^{i} = \sum_{h=0}^{H} c\left(\mathbf{s}_{t+h}^{i}, \mathbf{a}_{t+h}^{i}\right),
\end{equation}
where \( J^{i} \) denotes the total cost of the \( i \)-th trajectory. The control action \( \mathbf{a}_t \) is selected as a weighted average of the first actions in each trajectory.
\section{Approach: \texttt{Sym2Real}} 

Our method bridges the domain gap \emph{without} labor-intensive simulator tuning or hand-crafted physics priors, while remaining highly data-efficient. Our central idea is to first capture the \emph{core} governing dynamics of a system in the simplest possible setting, and then adapt it in the target environment with minimal target data. 
As shown in Fig.~\ref{fig:method}, this process unfolds in two stages: (1) learning a SR base model in a deliberately low-fidelity simulator, and (2) refining it through a few-shot residual learning with a lightweight NN in the target environment. 
Unlike prior approaches, this requires no explicit analytical equations; only a standard robot description file (e.g., MJCF) and a generic physics engine. 
Each step is detailed in the following subsections.

\subsection{Step I: Symbolic Regression in Low-Fidelity Simulation} \label{basefit}

The first stage distills the core system dynamics without simulation engineering and explicit system knowledge.

    \subsubsection{Low-fidelity simulation} Here, \emph{low-fidelity} means no manually injected real-world effects or system identification: no parameter tuning for payload, center of mass, or other hardware-specific properties, and no domain randomization \cite{tobin2017domain}. The simulator uses default robot description files, noiseless observations, instantaneous control effects, and lacks external disturbances added to the system.
    These simplifications remove the need to incorporate many of the typical robustness techniques for sim2real transfer \cite{11017653, gronauer2022using}, keeping the simulation setup lightweight and free from expert intervention.

    \subsubsection{SR base model fitting} 
    Given this environment, we follow an iterative trajectory optimization framework outlined in Section \ref{preliminary} to fit an SR model one episode at a time. 
    Each episode is capped at 400 environment steps. 
    We fit $D$ independent symbolic equations, one per state dimension. Each $f_{\text{SR},i}$ maps the complete state-action vector to the $i$-th component of the next state, so each next-state component can depend on any current state or action dimension:
    \begin{equation}
        [\mathbf{s}_{t+1}]_i = f_{\text{SR},i}(\mathbf{s}_t, \mathbf{a}_t), \quad \text{for } i = 1, \dots, D
    \end{equation}
    We use the open-source symbolic regression package PySR \cite{cranmer2023interpretable}, allowing the operators $\{+,\, -,\, \times,\, \cos,\, \sin\}$, capping the expression length at 85, running only 5 search iterations, and always warm starting from the previous iteration, which keeps the equations compact and the fitting process fast. The model is fitted with an $\ell_1$ loss. 
    These settings follow PySR's official tuning guidelines and involve straightforward adjustment of the parameters.
    In practice, we find that any maximum expression length long enough to capture the dominant dynamics yields comparable performance.

    \subsubsection{Safe Exploration of Risky Behaviors} Although episodes are truncated early upon contact with the ground, other dangerous trajectories such as continuous flipping or high-speed drifting are retained. These movements would have triggered safety mechanisms in the real world to protect the hardware, but they help span a wider range of state-action pairs, making it easier for the model to capture the underlying dynamics. We leverage the safety of simulation to include them for training.

\subsection{Step II: Adaptive Symbolic Regression Model}

Under domain shifts, our minimum performance assumption is that the base SR dynamics model is accurate enough to safely collect real-world data in the target domain without causing hardware damage. While this provides sufficient performance for safe exploration, it is rarely optimal in the target domains, as the initial controller stage is designed to capture only the underlying core equations. Thus, the goal in this step is to learn the residual with minimal physical interactions to correct these discrepancies and specialize the model to the target setting. 

The residual process starts with collecting an episode of target domain data with the frozen SR base model. Given a ground truth state-action pair \((\mathbf{s}_t, \mathbf{a}_t)\) and the next state, $\mathbf{s}_{t+1}^{\text{target}}$, the SR model produces an intermediate prediction \(f_{\text{SR}}(\mathbf{s}_t, \mathbf{a}_t)\). We augment this with the residual, $f_{\text{res}}(\mathbf{s}_t, \mathbf{a}_t)$, to get the next state, $\hat{\mathbf{s}}_{t+1}$:
\begin{equation}
    \hat{\mathbf{s}}_{t+1} = f_{\text{SR}}(\mathbf{s}_t, \mathbf{a}_t) + f_{\text{res}}(\mathbf{s}_t, \mathbf{a}_t).
\end{equation}

To learn the residual, we employ a multilayer perceptron (MLP), which is optimized to minimize the following loss:
\begin{equation}
    \mathcal{L} = 
    \underbrace{\left\| \mathbf{s}_{t+1}^{\text{target}} - \hat{\mathbf{s}}_{t+1} \right\|_2^2}_{\ell_2\text{ prediction loss}}
    +
    \underbrace{\lambda \left\| f_{\text{res}}(\mathbf{s}_t, \mathbf{a}_t) \right\|_2^2}_{\text{regularization}},
\end{equation}
where the first term is a standard $\ell_2$ loss and the second term serves as a regularization that encourages the residual output to remain small. 
The $\lambda$ coefficient tunes the overall reliance on the base SR model. 
We set it to keep the residual subordinate to the SR prediction to avoid overfitting to sparse target data.
The iterative process continues with collecting one additional trajectory. 
This few-shot residual process terminates when the model shows no meaningful improvement even with more data.

\subsection{Motivation for Controller Choice}

A key advantage of our framework is that it requires no simulation environment tuning, which is closely related to the choice of controller. 
Without domain randomization, reinforcement learning learns a policy tightly coupled to the simulated dynamics and fails under model mismatch, a behavior also observed in prior work \cite{kunapuli2025leveling}. 
In contrast, MPC optimizes actions online and tolerates moderate modeling errors. 
We validate this empirically: an RL policy trained on our simplified simulation did not transfer, whereas the model-based MPC controller allowed the drone to hover safely.

\subsection{Implementation Details}
MuJoCo was used as the physics simulator. Inspired by prior work on sampling-based MPC \cite{Turrisi_2024}, we implemented both the dynamics models and MPC control scripts in JAX to support efficient batched rollouts for real-time control. All simulation experiments ran on a single NVIDIA L40 GPU, and real-world experiments ran on an RTX 3090.
\section{Experiment Setup}

We evaluated our method in both simulation and physical settings on two robotic platforms, \textit{Crazyflie 2.1} quadrotor and \textit{MuSHR} racecar \cite{srinivasa2019mushr}, using the following setup and metrics.
        
    \subsection{System Platforms}
        \noindent\textbf{Crazyflie 2.1 quadrotor} The quadrotor is described with a 13-dimensional state vector \( \mathbf{s}_t \in \mathbb{R}^{13} \), consisting of global position \( \mathbf{p}_t = (x, y, z)_t \in \mathbb{R}^3 \), orientation \( \mathbf{q}_t = (q_w, q_x, q_y, q_z)_t \in \mathbb{R}^4 \) represented as a unit quaternion, global linear velocity \( \mathbf{v}_t = (\dot{x}, \dot{y}, \dot{z})_t \in \mathbb{R}^3 \), and body angular velocity \( \boldsymbol{\omega}_t = (\omega_x, \omega_y, \omega_z)_t \in \mathbb{R}^3 \). 
        The control input \( \mathbf{a}_t = (a_0, a_1, a_2, a_3)_t \in \mathbb{R}^4 \) consists of the total thrust ($a_0$) and desired angular velocity components $(a_1, a_2, a_3)$, tracked by the onboard rate controller; this interface is low-level enough to expose the force and rotational dynamics to the model while remaining stable at offboard command rates.
        Predicted quaternions are normalized at each rollout step. 
        
        \noindent\textbf{MuSHR racecar} The racecar state is represented by a 7-dimensional vector that consists of position, \(\mathbf{p}_t = (x, y)_t\), yaw orientation \(\mathbf{R}_t = (\sin\theta_{yaw}, \cos\theta_{yaw})_t\), linear velocity \(\mathbf{v}_t = (v_x, v_y)_t\), and the yaw rate \(\boldsymbol{\omega}_t = (\omega_{yaw})_t\). The action \(\mathbf{a}_t = (a_0, a_1) \in \mathbb{R}^2\) consists of the steering angle and the commanded longitudinal velocity.

    \subsection{Real-World Hardware Configuration} 
    \label{sec:hardware-setup}
    External motion capture (MoCap) data was streamed at 100 Hz to provide pose measurements. Crazyflie used only position from MoCap, while orientation and angular rates were from the onboard IMU. MuSHR racecar used both from MoCap. 
    The velocities were computed from consecutive measurements. All computation was performed offboard. Control commands were transmitted at 50 Hz for the Crazyflie via Crazyradio 2.0 and at 20 Hz for the MuSHR racecar via Wi-Fi through the MuSHR ROS stack. 
    
    For Crazyflie, thrust representation posed a unique issue. On the hardware, thrust was represented as a 16-bit unsigned integer ($0$-$65535$), requiring conversion from the Newton units used in simulation. While an exact mapping would require calibration with a load cell, we applied an approximate linear mapping based on the values hard-coded in the firmware. 
    Our method was designed to mitigate inaccuracies from this, as well as unmodeled communication latency.

    \subsection{Task Definition and Performance Evaluation Metrics}
    All tasks are defined as tracking the target state $\mathbf{s}^*$ along a trajectory. The MPC cost over the horizon $H$ is defined as:
    \begin{multline}
        \label{eq:cost}
        J = \sum_{k=0}^{H-1} \Big(
            \underbrace{\alpha_p \left\| \mathbf{p}_k - \mathbf{p}^* \right\|_2}_{\text{position cost}} +
            \underbrace{\alpha_r \, d_R(\mathbf{R}_k, \mathbf{R}^*)}_{\text{orientation cost}} + {} \\
            \underbrace{\alpha_v \left\| \mathbf{v}_k - \mathbf{v}^* \right\|_2}_{\text{velocity cost}} +
            \underbrace{\alpha_\omega \left\| \boldsymbol{\omega}_k - \boldsymbol{\omega}^* \right\|_2}_{\text{angular velocity cost}}
        \Big)
    \end{multline}
    where 
    \( d_R(\mathbf{R}_k, \mathbf{R}^*) = 1 - (\mathbf{q}_k \cdot \mathbf{q}^*)^2 \)
    for the Crazyflie, and 
    \(
    d_R(\mathbf{R}_k, \mathbf{R}^*) = \left\| \sin(\theta_k) - \sin(\theta^*) \right\|_2 + \left\| \cos(\theta_k) - \cos(\theta^*) \right\|_2
    \)
    for the MuSHR. 
    The coefficients \( \alpha_p, \alpha_r, \alpha_v, \alpha_\omega \) are weights for each cost term; they were tuned in simulation and held fixed across all sim2sim and sim2real adaptation phases.
    For the performance metric, we use the Euclidean position error between the current position and the reference position, 
    \( e_t = \lVert \mathbf{p}_t - \mathbf{p}^* \rVert_2 \),
    averaged over the entire trajectory length. 

    \begin{figure}[t]
    \vspace{5pt}
    \centering
    \includegraphics[width=0.95\linewidth]{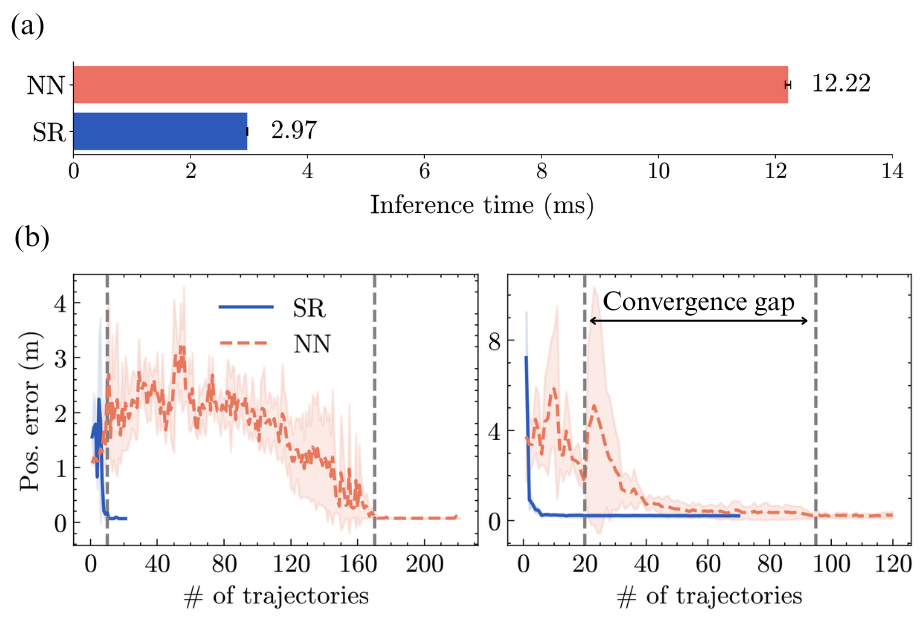}
    \caption{\textbf{SR vs. NN base model efficiency.} (a) A single step model inference speed of the base models. (b) The data efficiency of the SR model in the simplified simulation environment.}
    \label{fig:from-scratch}
    \vspace{-5pt}
\end{figure}
\begin{table}[t]
\vspace{5pt}
\centering
\setlength{\tabcolsep}{4pt}
\renewcommand{\arraystretch}{1.15}
\resizebox{1\columnwidth}{!}{%
\begin{threeparttable}
\caption{Representative equations from SR dynamics model.}
\label{tab:sr_equations}

\begin{tabularx}{\columnwidth}{@{} l l >{\centering\arraybackslash}X @{}}
\toprule
System & Next state & Representative SR equation \\
\midrule
\multirow{6.5}{*}{\centering Crazyflie}
  & \multirow{3}{*}{$z'$ (position)}  & $z + 0.02\,(v_z - \sin(\sin(\cdot)))$ \\
  &                 & $+\,0.0087\underbracket{a_0}_{\text{thrust}}\sin(\cos(\cdot))$ \\
  \cmidrule(lr){2-3}
  & \multirow{3}{*}{$v_z'$ (lin. velocity)} & $v_z - 0.67$ \\
  &                   & $+\,(-1.65)\underbracket{a_0}_{\text{thrust}}q_y^{2} + \sin(\cos(\cdot)+\underbracket{a_0}_{\text{thrust}})$ \\
\midrule
\multirow{7}{*}[0.8ex]{\centering MuSHR}
  & \multirow{2}{*}{$x'$ (position)}     & $x - (-0.0469 \cdot( (-0.0647\overbracket{a_{1}}^{\text{vel.}}) \cdot (\cdot)$\\
  &                & $\;+\; (\cdot\underbracket{a_{0}}_{\text{ang.}} \cdot (\cdot)) \;+\; v_{x} ))$ \\
  \cmidrule(lr){2-3}
  & \multirow{2.25}{*}{$\omega_{yaw}'$ (angular vel.)} & $((\cdot)+0.233\underbracket{a_1}_{\text{vel.}})\sin(\sin(\underbracket{a_0}_{\text{ang.}}))$ \\
\bottomrule
\end{tabularx}

\begin{tablenotes}
\footnotesize
\item For visual clarity, $(\cdot)$ denotes truncated terms (constants and/or other variables).
\end{tablenotes}

\end{threeparttable}
}
\vspace{-15pt}
\end{table}

\section{Base Controller Experiments}

First, we tested the SR base model efficiency when trained from scratch as described in Section \ref{basefit}.
During data collection, the quadrotor was tasked to follow hovering, circular, and lemniscate paths, while the racecar was tasked to follow circular paths and park at a goal from random initial states. 
This diversity exposes the system to a wide range of states and actions, ensuring that measured data efficiency reflects the modeling approach rather than overfitting to a narrow set of behaviors.
Later, in Section~\ref{sec:adapation_stateaction}, we deliberately restricted state-action coverage to study how the models behave under limited coverage. We compared the SR model to an NN model as a baseline. Following the state-of-the-art model-based approach \cite{chua2018deep}, we implemented an ensemble of neural networks. Specifically, we used 4 ensembles of MLPs, each consisting of 4 layers with 200 hidden dimensions.

\begin{figure}[t]
    \vspace{5pt}
    \centering
    \includegraphics[width=\linewidth]{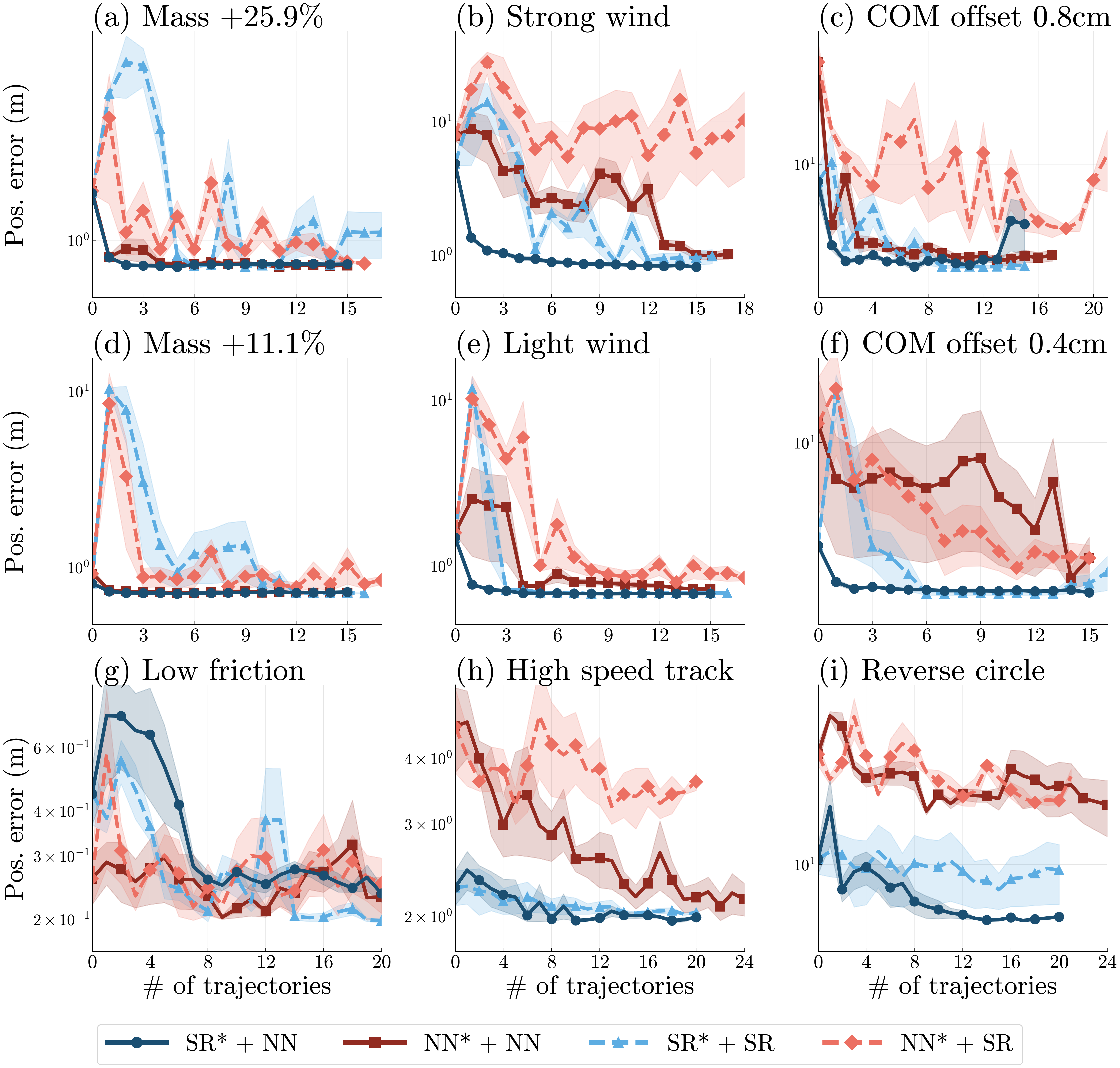}
    \caption{\textbf{Residual adaptation performance.} Episode cost during online adaptation for (a-f) the Crazyflie quadrotor under various disturbances and (g-i) the MuSHR racecar across scenarios. Results are averaged over three seeds, with shaded regions indicating standard deviation.}
    \label{fig:transfer-full-comparison}
    \vspace{-10pt}
\end{figure}
\begin{figure}[t]
    \includegraphics[width=\linewidth]
    {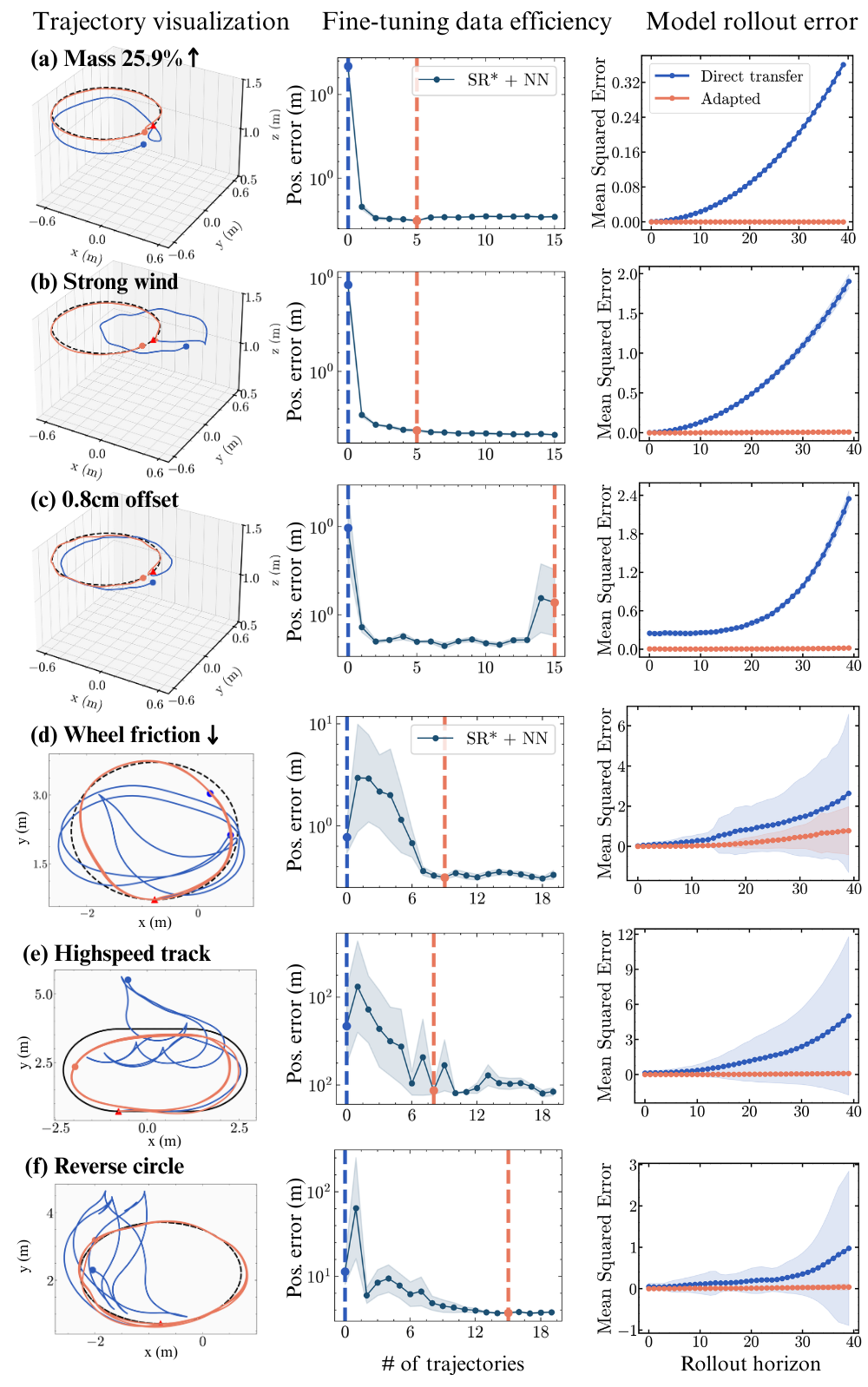}
    \caption{\textbf{Sim2sim adaptation performance of SR*+NN.} Few-shot residual adaptation improves both model prediction and control compared to direct SR transfer: (a-c) shows Crazyflie quadrotor adaptation and (d-f) shows MuSHR racecar adaptation. Dotted black lines represent target reference trajectories.
    }

    \label{fig:sim-adaptation-performance}
    \vspace{-15pt}
\end{figure}

\noindent\textbf{Efficiency} As shown in Fig.~\ref{fig:from-scratch}, the SR model demonstrated superior data efficiency compared to the NN baseline. 
For the racecar, it converged in fewer than 20 trajectories, whereas the NN model required over 100 trajectories. 
This advantage was even more pronounced in the higher-dimensional quadrotor, where the SR model achieved an accurate dynamics representation with just 10 trajectories, compared to approximately 160 required by the NN model. 
We measured efficiency with trajectory count rather than sample count to reflect real-world deployment practicality. 
Sampling-based MPC requires computationally efficient model rollouts that satisfy real-time control constraints. 
The compact nature of SR representation yielded action selection that was over 4 times faster compared to NN when evaluated with a quadrotor model and a controller with 1024 samples and a horizon of 0.8 seconds.
This computational efficiency can allow the evaluation of more candidate trajectories given the same amount of time, improving control performance. 

\noindent\textbf{Symbolically Regressed Dynamics Equations} Unlike standard SR benchmarks that focus primarily on minimizing prediction loss, robotic control requires fitted dynamics equations to reflect meaningful action-to-next-state dependencies.
This ensures the learned model captures the physics of how actions influence future states.
Without such dependencies, an MPC controller cannot properly evaluate candidate actions, since predicted states would remain unaffected by control inputs.
In our work, we explicitly examined whether symbolic models discovered by SR satisfy this requirement. For the quadrotor, we verified that thrust appears in the equations for both the next $z$-position and vertical velocity, capturing the direct influence of thrust actuation on vertical motion (Table~\ref{tab:sr_equations}). For the MuSHR racecar, we observed that both action variables, angular velocity and steering angle, appear in the position and heading equations. These representative cases demonstrate that the SR-fitted dynamics are physically plausible and suitable for control.
\section{Adaptation Experiments: Sim2Sim} 
In this section, we evaluate how efficiently our method can adapt to out-of-distribution (OOD) domain shifts in simulation. These include both the internal and external physics parameter changes and state-action coverage shifts.

\subsection{Adaptation Under OOD Dynamics Shift}

\noindent\textbf{Disturbances} For the quadrotor, we evaluated three categories of dynamics shifts: (i) internal mass changes (mass increased by $11.1\%$ and $25.9\%$ from the default 27g), (ii) structural changes via center-of-mass offsets along the $x$-axis, and (iii) external disturbances via planar wind applied along both $x$ and $y$ directions. For the racecar, we tested with lowered friction between the floor and the wheels, which introduced drifting behavior.

\noindent\textbf{Baselines} To compare the decomposed model design, we evaluated four combinations of base + residual model: 1) SR* + NN, 2) SR* + SR, 3) NN* + NN, and 4) NN* + SR, where * denotes a frozen base model. 

\noindent\textbf{Results} 
Fig. \ref{fig:transfer-full-comparison} shows adaptation efficiency and performance across the baselines. 
Although the performance varies across different scenarios, SR* + NN showed the most data-efficient and stable performance across the dynamics shifts. 
In mass and offset shift scenarios, SR*+NN started at a lower cost upon zero-shot deployment and improved the quickest. 
Under weak wind, the initial performance was similar, but the SR* base models converged to a lower episode cost, and for the strong wind, it started at a higher cost but converged to a lower cost while NN* base variants struggled to converge. 
Overall, SR performed well as a base core extractor and NN learned the residuals efficiently. 
\begin{table}[t]
    \vspace{5pt}
    \centering
    \setlength{\tabcolsep}{6pt}
    \renewcommand{\arraystretch}{1.1}
    \caption{
    Zero-shot success rate (\%) under OOD state–action.
    coverage. 
    Each entry is averaged over 20 trajectories.
    }
    \label{tab:sa-ood-zeroshot}
    \begin{tabular}{l c c|c c}
        \toprule
        \multirow{2}{*}{Model} & 
        \multicolumn{2}{c|}{Racecar (OOD start)} & 
        \multicolumn{2}{c}{Quadrotor (unseen goal)} \\
        \cmidrule(lr){2-3} \cmidrule(lr){4-5}
        & Train & Test & Train & Test \\
        \midrule
        SR base  & \multicolumn{1}{c}{100\%} & \multicolumn{1}{c}{90\%}  & \multicolumn{1}{c}{100\%} & \multicolumn{1}{c}{100\%} \\
        NN base & \multicolumn{1}{c}{100\%} & \multicolumn{1}{c}{20\%}  & \multicolumn{1}{c}{100\%} & \multicolumn{1}{c}{30\%} \\
        \bottomrule
    \end{tabular}
    \vspace{-10pt}
\end{table}
\begin{figure}[t]
    \centering
    \includegraphics[width=0.87\linewidth]{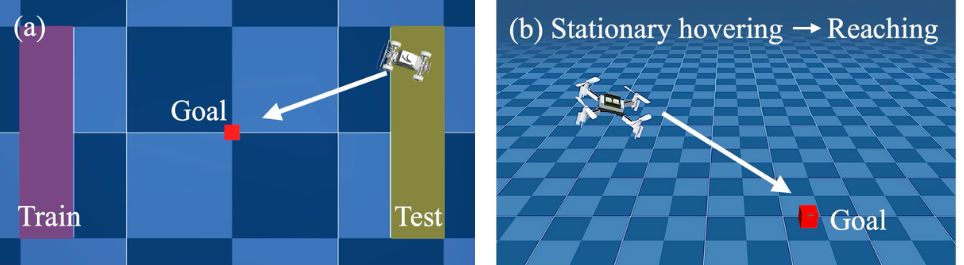}
    \caption{\textbf{Experiment setup for zero-shot generalization under OOD state–action coverage.} (a) The MuSHR racecar was trained with initial positions restricted to a narrow range (purple), and tested from disjoint start states (yellow). (b) The quadrotor was trained to remain hovering at a fixed location (red cube), but at test time, it was initialized from a new goal and asked to fly and hover at the red cube.}
    \label{fig:sa-ood-setup}
    \vspace{-10pt}
\end{figure}
\begin{figure*}[t]
    \centering
    \includegraphics[width=\linewidth]{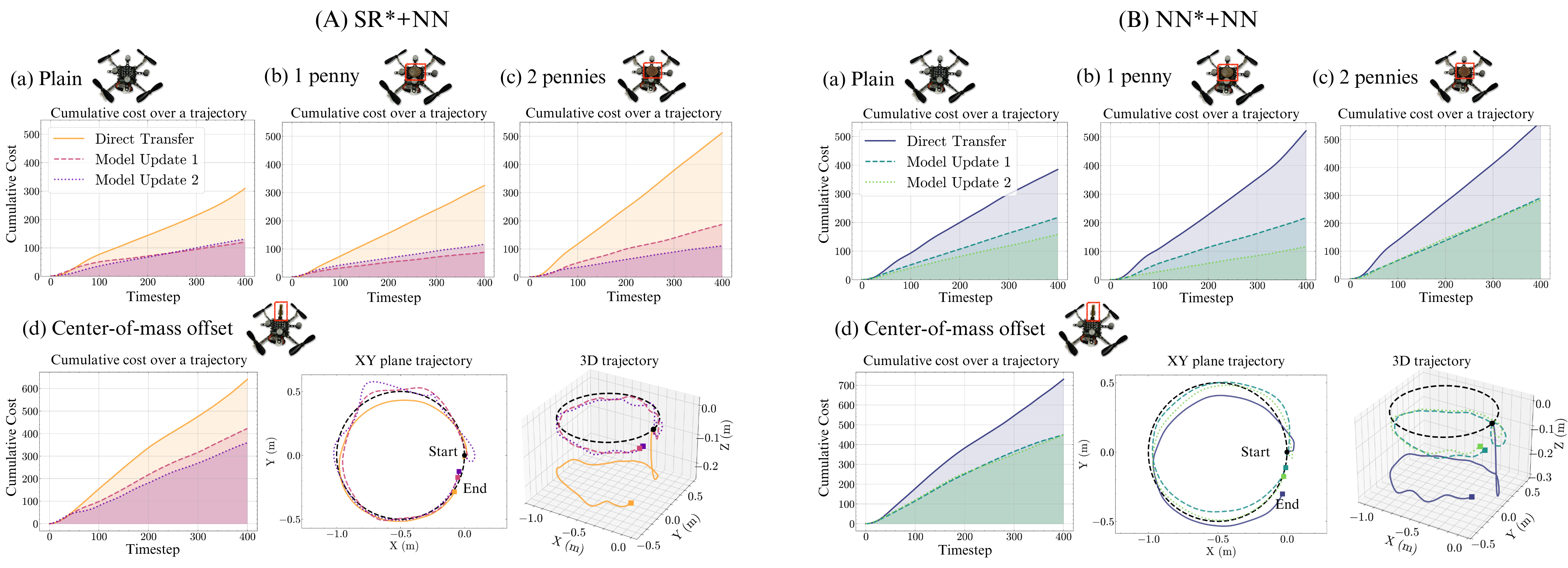}
    \caption{\textbf{Sim2real Crazyflie 2.1 quadrotor adaptation.} (a) Plain sim2real transfer, (b) added mass (one penny), (c) added mass (two pennies), and (d) center-of-mass offset. In (a) and (b), the quadrotor was tasked with hovering, while in (d) it was tasked to follow a circular trajectory.}
    \label{fig:rw-exp-cf}
    \vspace{-12pt}
\end{figure*}

\begin{figure}[t]
    \centering
    \includegraphics[width=\linewidth]{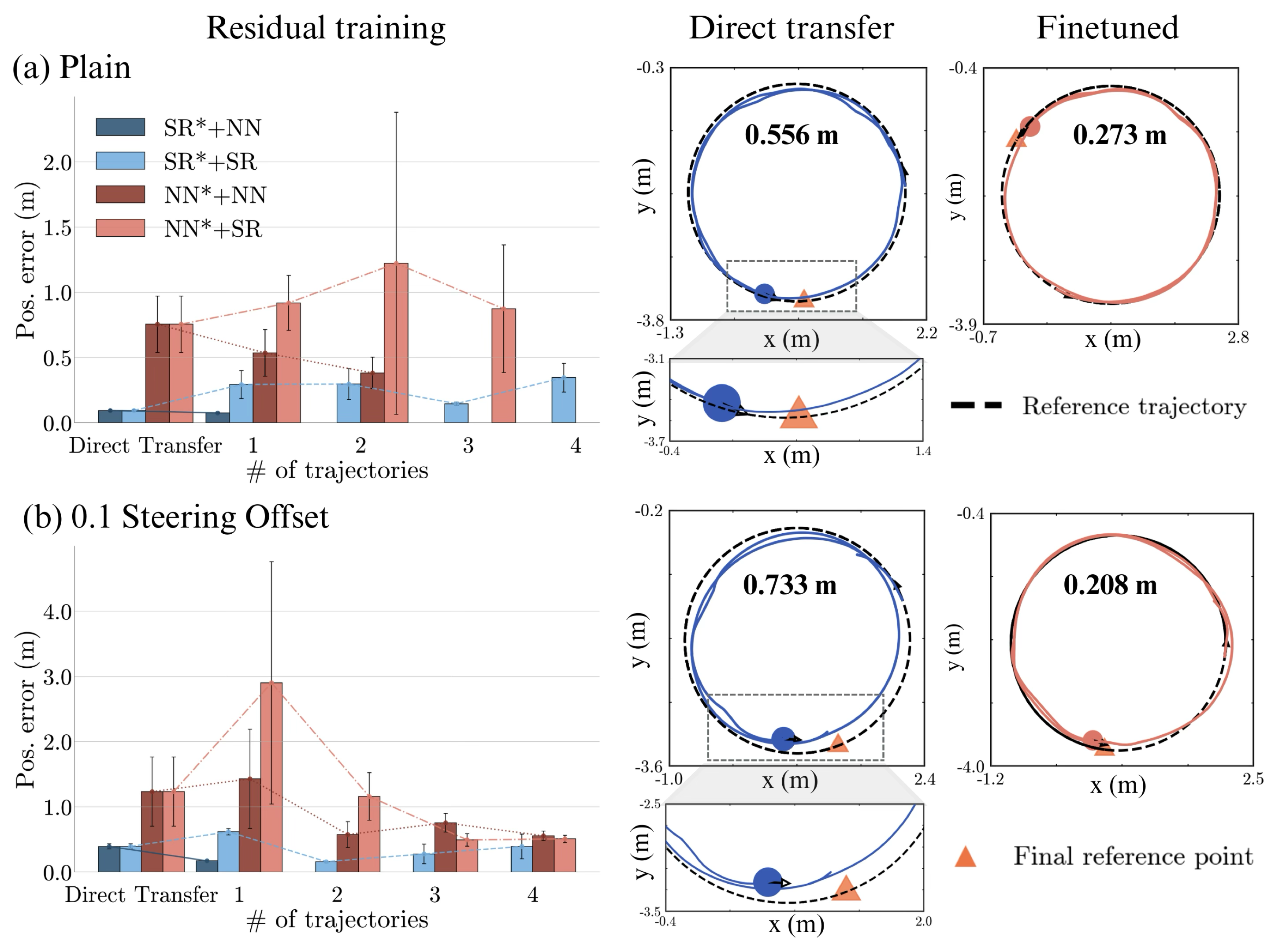}
    \caption{\textbf{Sim2real MuSHR racecar adaptation.} (a) Plain sim2real transfer and (b) steering-offset setting. Performance of all baselines and \texttt{Sym2Real} (i.e., SR* + NN) across trials of few-shot adaptation, averaged over three runs is reported. SR* + NN residual adaptation improved performance, illustrated with a sample trajectory compared with direct SR transfer.}
    \label{fig:rw-exp-mushr}
    \vspace{-10pt}
\end{figure}

Building on this finding of SR* + NN's ability for efficient adaptation, we further visualized the performance of the SR* + NN model after few-shot learning compared to the initial SR model. 
Fig.~\ref{fig:sim-adaptation-performance} shows that the SR* + NN residual model quickly improves both the model and the control performance with fewer than 10 additional trajectories across all 6 OOD scenarios, demonstrating robust few-shot adaptation with limited data.

\subsection{Adaptation Under OOD State-Action Coverage} \label{sec:adapation_stateaction}

Additionally, we experimented with generalization when the training coverage was deliberately restricted. 
It is both impractical and time-consuming to provide training data that represents the entire state–action space. We therefore evaluated how models behave under limited coverage.

\noindent\textbf{Experiment setup} 
Fig. \ref{fig:sa-ood-setup} shows the training setup. For the racecar, the goal was to park at $(x=1, y=0)$ with initial positions sampled from $x \in [-1.0, -0.5]$ and $y \in [-1.0, 1.0]$, and yaw randomized between $-45^\circ$ and $45^\circ$.
For the quadrotor, training task was hovering at a fixed location with an upright orientation and zero initial velocities. At test time, the racecar was initialized from an unseen start range of $x \in [2.5, 3.0]$, $y \in [-1.0, 1.0]$. This setup naturally encouraged forward motion during training but required backward driving at test time. 
For the quadrotor, goals were randomized to a $0.6m \times 0.6m \times 1m$ volume. 

\noindent\textbf{Results} 
Table~\ref{tab:sa-ood-zeroshot} shows that SR models recovered the underlying dynamics and succeeded under this shift with $100\%$ success rate, while NN models overfit to the training range and failed to adapt. 
Both SR and NN models struggled more with racecar state-action shifts ($90\%$ and $20\%$ respectively), so we further adapted through residual learning to assess adaptation efficiency. 
We tested high-speed ($\times 4$) tracking on a new unseen track and tracking reverse circular paths. Fig.~\ref{fig:transfer-full-comparison} shows SR variants adapted much faster than NN baselines, converging with fewer trajectories.
The results demonstrate that SR models captured global dynamics structure, enabling robust zero-shot generalization under distribution shifts. NNs overfit unless training explicitly covers test regimes. SR base variants adapted faster since they already extracted compact and representative dynamics, requiring fewer corrections.
\section{Adaptation Experiments: Sim2Real}

We further tested our method on sim2real adaptation cases. 

\noindent\textbf{Disturbances} For the Crazyflie 2.1, we first considered the natural sim2real gap (plain), which arises from factors such as Vicon markers, state estimation noise, delays in communication and motor actuation, and oversimplification of thrust conversion as discussed in Section \ref{sec:hardware-setup}. 
To further increase the gap, we introduced additional disturbances by attaching $2.5g$ pennies at the center of mass and adding a $3g$ metal weight offset from the center of mass to create structural imbalance. 
For the racecar, we tested the performance under natural sim2real gap and steering offset (+0.1).

\noindent\textbf{Results} A key prerequisite for sim2real adaptation is that the transferred model produces predictions accurate enough for
safe real-world data collection.
As shown in Fig~\ref{fig:rw-exp-cf} and~\ref{fig:rw-exp-mushr}, although there was a mismatch between the reference and the zero-shot SR model trajectories, the SR model was accurate enough for the real-world deployment, which is especially crucial for passively unstable systems like quadrotors.

Fig.~\ref{fig:rw-exp-cf} illustrates the Crazyflie experiments. Across all disturbance scenarios, residual learning consistently improved control performance, leading to trajectories that more closely tracked the reference. 
The benefit was especially evident in the circular path tracking task under an offset shift: the residual-augmented controller improved the alignment with the reference path. At the end of the cycle, the trajectory returned to a position much closer to the starting point, demonstrating tighter long-horizon tracking. Both SR*+NN and NN*+NN showed stable control improvement over few-shot residual training, showing comparable adaptation performance.
The SR residual baselines exhibited unstable performance fluctuations during adaptation that resulted in hardware damage. Lastly, when we directly fine-tuned the SR model with real-world data (i.e., SR*+SR), we observed the SR model losing key structural dependencies identified in simulation. In particular, the thrust variable, which correctly appears in the $z$ velocity update in Table~\ref{tab:sr_equations}, was absent in the newly fitted equation. This omission led to inaccurate 
vertical dynamics and failure
to maintain stable flight. 
Fig.~\ref{fig:rw-exp-mushr} shows the experimental results for the racecar. 
In both scenarios, the SR base model showed superior zero-shot performance compared to the NN base model, and this gap is more noticeable than in sim2sim scenarios. 
Additionally, SR*+NN showed consistent improvement over both cases, whereas SR*+SR produced less stable improvements. 

\section{Conclusions, Limitations, and Future Work}

We propose \texttt{Sym2Real}, a data-efficient method for acquiring robust low-level models for a quadrotor and a racecar for initial control and further OOD adaptation. 
It requires minimal simulation tuning, data collection, and expert knowledge.  
A key limitation is that SR scales poorly as state dimensionality grows, restricting our method to relatively low-DoF systems; 
extending it to higher-DoF robots such as manipulators or legged platforms is a promising direction, potentially via sparse NNs or methods for discovering low-dimensional structure from high-dimensional data \cite{huang2026automated}. 
Decomposing dynamics into symbolically regressable and non-regressable components could address contact rich systems.





\bibliographystyle{IEEEtran}
\bibliography{
    references/intro,
    references/rw_partA, 
    references/rw_partB, 
    references/closest_work,
    references/others
}
\end{document}